\documentclass{article}
\usepackage{spconf,amsmath,graphicx}

\usepackage{xcolor, soul}
\sethlcolor{green}
\usepackage{hyperref}

\title{Spoken Language Biomarkers for Detecting Cognitive Impairment}

\name{Tuka Alhanai\sthanks{The authors would like to thank Maggie Sandoval from Evidation Health Inc., and the team at Boston University and The Framingham Heart Study: Ida Xu, Brynna Wasserman, Maulika Kohli, Nancy Heard-Costa, Yulin Liu, Karen Mutalik, Mia Lavallee, Alvin Ang, and Spencer Hardy, for their efforts in curating and processing the data for this study. Tuka would like to acknowledge the Abu Dhabi Education Council for sponsoring her graduate studies. Contact: \texttt{\{tuka,glass\}@mit.edu, rhodaau@bu.edu} }$^\dagger$, Rhoda Au$^{\ddagger\mathsection}$, and James Glass$^\dagger$ }
\address{$^\dagger$Computer Science and Artificial Intelligence Laboratory, 
		\\ Massachusetts Institute of Technology, Cambridge MA, USA \\
		$^\ddagger$Departments of Anatomy \& Neurobiology, Neurology, and Epidemiology, \\ Boston University School of Medicine and Public Health, Boston MA, USA \\
        $^\mathsection$The Framingham Heart Study, Framingham MA, USA\\ }
\begin{document}
%
\maketitle

\begin{abstract} 
In this study we developed an automated system that evaluates speech and language features from audio recordings of neuropsychological examinations of 92 subjects in the Framingham Heart Study. A total of 265 features were used in an elastic-net regularized binomial logistic regression model to classify the presence of cognitive impairment, and to select the most predictive features. We compared performance with a demographic model from 6,258 subjects in the greater study cohort (0.79 AUC), and found that a system that incorporated both audio and text features performed the best (0.92 AUC), with a True Positive Rate of 29\% (at 0\% False Positive Rate) and a good model fit (Hosmer-Lemeshow test $>$ 0.05). We also found that decreasing pitch and jitter, shorter segments of speech, and responses phrased as questions were positively associated with cognitive impairment. 

\end{abstract}


\section{Introduction} \label{sec:intro}

Cognitive impairment is a decline in mental abilities that is severe enough to interfere with daily life \cite{nussbaum2003alzheimer}. Dementia is a particularly debilitating type of cognitive impairment costing \$200 billion in the USA alone \cite{prince2011world, leifer2003early, alzheimer20152015}, and coming second only to spinal-cord injuries and terminal cancer in the severity of its effects \cite{world2003world, ferri2006global}. The World Health Organization (WHO) reports that early diagnosis of dementia provides a pathway to treatments and an opportunity to plan ahead \cite{prince2011world}. 


Studies of cognitive impairment and dementia (Alzheimer's Disease, Vascular Dementia, Lewy Bodies Dementia, etc.) have explored multiple modalities of information for assessment and diagnosis. This includes subjective measures of cognitive decline (e.g., patient's response to the question `Has your memory become worse?') \cite{saykin2006older, jessen2010prediction, reisberg2010outcome}, medical profile (stroke, cardiovascular disease, blood pressure, etc.) \cite{newman2005dementia,satizabal2016incidence}, education level \cite{ott1995prevalence, cobb1995effect}, imaging exams \cite{au2006association,jack2009serial,mosconi2008multicenter}, apolipoprotein E (APOE) genotype (from plasma samples) \cite{farrer1997effects,kim2009role,myers1996apolipoprotein}, atherosclerosis (via ultrasonography) \cite{hofman1997atherosclerosis}, brain-derived neurotrophic factor (BDNF) \cite{weinstein2014serum}, cerebro-spinal fluid \cite{shaw2009cerebrospinal}, and other laboratory measures (glucose homeostasis, markers of inflammation, blood homocysteine, folate, vitamin B-12, etc.) \cite{quadri2004homocysteine,van2012biomarkers}.

While these studies have explored associations between their measures and cognitive outcome, such information has a high barrier of acquisition due to the costly nature of laboratory tests and imaging scans. This motivates the exploration of measures that are easier to record and are less invasive, specifically, speech. 
Studies using speech and language measures have modeled a subject's language by capturing patterns of word usage (Part-of-Speech tag, vocabulary to total words ratio, etc.)~\cite{ripich1991turn, thomas2005auto}, and conversation acts (response to question, active listening, etc.)~\cite{lopez2015automatic}. Other studies have used acoustic information from the speech waveform including speech/silence segments, onset time of speech, time intervals between words in a sentence as aligned with a transcript \cite{konig2015automatic}, as well as voice quality measures (pitch, shimmer, jitter, etc.) \cite{meilan2014speech, lopez2015automatic}. Speech data used in experiments varied from being highly structured memory and reading tests \cite{meilan2014speech, konig2015automatic, fraser2016linguistic}, to organic, open-ended, conversation-like interactions (e.g. sharing stories) \cite{atay2015automated,lopez2015automatic}.


\section{OBJECTIVES}

Our objectives are to (1) extract and identify audio and text features that are most predictive of cognitive impairment, and (2) to thoroughly compare the statical performance of a predictive model using the identified features against established baselines \cite{satizabal2016incidence}. 

Our study differentiates itself by exploring an extensive set of features covering both linguistic (text) and acoustic information, while also accounting for the demographics (age, gender, etc.) of subjects. Our study processed the world's largest volume of audio/text data on subjects with cognitive impairment to date (100 hours), and utilized audio recorded during neuropsychological examinations, without explicitly modeling the underlying structure of the interaction. This allows for the methods developed to generalize to other contexts in which an individual engages in spoken interactions.  

\section{Methods} \label{sec:methods}


\subsection{Data}

The data used in this study was collected from the Framingham Heart Study, an on-going longitudinal population study of 15,447 subjects from 1948 to the present \cite{mahmood2014framingham}. 
Since 1999 most subjects have undergone neuropsychological examinations \cite{satizabal2016incidence}, and as of 2005, it became standard to record audio of these examinations. The neuropsychological examinations include multiple components to assess memory, attention, executive function, language, reasoning, visuoperceptual skills, and premorbid intelligence. All participants provided written informed consent, with study protocols and consent forms approved by the institutional review board at the Boston University Medical Center. 

Our study used 92 audio recordings of neuropsychological examinations that had available text transcripts. Recordings were on average, 65 minutes in duration, contained 2,496 words, with a vocabulary size of 527 words. We also included 6,258 subjects (those that had no missing data) with the same set of demographic variables as the 92 subjects with audio and text transcripts. Additional information on subject profiles is provided in Table \ref{table:dataset}.

Transcripts for each audio file were generated manually using rules developed in~\cite{glass2004analysis}. Transcribers were instructed to include timestamps for each speaker turn (subject/tester), indicate who spoke when, transcribe speech orthographically (e.g. nineteen dollars instead of \$19), include tags to highlight moments such as filled pauses ($<$um$>$), and to subjectively insert punctuation.


\begin{table}[h!]
\footnotesize
\centering
\caption{\textbf{Dataset Statistics.} {\small Subjects with and without audio recordings.}}
\label{table:dataset}
\begin{tabular}{lcc}
& \\
\hline \hline

& \begin{tabular}[c]{@{}c@{}} Audio \end{tabular}    
& \begin{tabular}[c]{@{}c@{}} Demographic \end{tabular} \\ \hline

\textbf{Subjects} 					\hfill $N$			   & 92            & 6,258 \\
\textbf{Examples} 					\hfill total           & 92            & 12,258* \\
\textbf{Outcome} 					\hfill Impaired (\%)   & 21 (22.8)     & 928 (7.6) \\
\textbf{Age}, years					\hfill mean (sd)       & 68 (17)       & 66 (15) \\
\textbf{Gender} 					\hfill male (\%)       & 47 (51)       & 5,349 (44) \\
\textbf{Duration}, minutes			\hfill mean (sd) 	   & 65 (18)       & - \\
\textbf{Duration}, hours			\hfill total      	   & 100 	       & - \\
\textbf{Vocabulary Size}, words		\hfill mean (sd) 	   & 527 (181)     & - \\ 
\textbf{Transcript Size}, words		\hfill mean (sd) 	   & 2,496 (1,508) & - \\ \hline
\multicolumn{3}{l}{*Some subjects had multiple neuropsychological examinations.} \\
\end{tabular}
\end{table}
\subsection{Outcome of Interest}

Our outcome of interest was a binary indicator of cognitive impairment, with impairment coded as 1. We labeled subjects as cognitively impaired if the date of impairment (as concluded by the dementia diagnostic review panel \cite{seshadri2006lifetime}) was on or before the date of the neuropsychological examination where the audio recording took place. Using this criteria, 21 subjects (22.8\%) were cognitively impaired. Ten of these subjects had a severity rating less than mild, six were mild, five were moderate, and none were severe \cite{seshadri2006lifetime}. Fourteen subject were diagnosed as having Alzheimer's disease using the NINCDS-ADRDA criteria \cite{mckhann2011diagnosis}, and five were diagnosed with Vascular dementia based on the NINCDS-AIRENS criteria \cite{roman1993vascular}. 

Since the objective of this study was to detect cognitive impairment, we did not differentiate between subjects with varying levels of severity, nor underlying conditions (Alzheimer's Disease, Vascular Dementia). 

\subsection{Features}
We extracted a total of 265 features of three types: 14 demographic, 230 acoustic, and 21 text\footnote{Code: \scriptsize{\url{https://github.com/talhanai/asru2017-method.git}} }. All continuous features were z-scored, and all categorical features were dummy coded.  

\textbf{Demographic features} contained the subject's age, sex, highest level of self-reported education (didn't graduate high school, high school graduate, attended but didn't graduate college, or college graduate or higher), and occupation (part-time, full-time, not working due to disability, retired, unemployed, never worked, volunteer, full-time student, or other). Age was modeled as a continuous feature, while all other features were dummy coded to represent categories, resulting in 14 features total.

\textbf{Acoustic Features} were extracted using the openSMILE v2.1 toolkit over 20ms frames, shifted 10ms, on audio files that were downsampled to 8kHz \cite{eyben2010opensmile}. Features contained information on the subject's pitch, probability of voicing, Root-Mean-Square (RMS) energy, Mel Frequency Cepstral Coefficients (MFCCs), Harmonic to Noise Ratio (HNR), zero crossing rate, shimmer, and jitter, as well as the difference between each features in neighboring frames. This resulted in 46 frame-level features, the details of their calculation are described by \textit{Eyben} \cite{eyben2015real}. 
We were interested in capturing high-level statistics from frames that were most likely to be speech (as opposed to non-speech frames), and processed the features to generate the mean, maximum, minimum, median, and standard deviation. This resulted in 230 global features that describe each subject's speech characteristics over the entire exam. The specific steps to compute the global features from the frame-level features are as follows: 

\begin{enumerate}  \itemsep0em 
\item \textbf{Feature Normalization:} To remove information of the recording environment, all features except probability of voicing, pitch, shimmer, and jitter (where the absolute number matters) were z-scored across each subject.
\item \textbf{Speaker Turn:} Using the speaker turn labels and timestamps in the transcript, we processed the frames that belonged to the subject only, and not the tester.
\item \textbf{Normalization:} We z-scored the probability of voicing per subject. (This was only used to extract speech segments, the unnormalized probability of voicing was used to calculate the global feature.)
\item \textbf{Smoothing:} We calculated an envelope over the z-scored probability of voicing by a peak (upper) envelope using a spline over local maxima separated by at least 10 points (100ms).
\item \textbf{Speech Segments:} We labeled frames as speech if the threshold of the smoothed probability of voicing was greater than 0.1 standard deviations from the (zero) mean.
\item \textbf{Global Features:} We calculated the mean, maximum, minimum, median, and standard deviation of these frame-level features (from the speech segments) for each subject.
\item \textbf{Non-zero Pitch:} For the pitch, shimmer, and jitter, we calculated the same global features, but for non-zero values (i.e. presence of pitch activity) within the speech segments.
\end{enumerate}

\textbf{Text features} contained the subject's number of words, duration of speech, speaking rate, questions, hesitations, vocabulary, and language perplexity (how well can the subject's next word be predicted). A total of 21 features were generated, as follows:

\begin{itemize} \itemsep0em
\item \textbf{Number of Words} (5 features): Number of Words were calculated for each turn the subject spoke, then the mean, minimum, maximum, median, and sum were calculated across all segments for each subject.\\
\item \textbf{Duration} (4 features): Duration was calculated for each turn the subject spoke, then the mean, minimum, maximum, and median were calculated across all segments for each subject.\\
\item \textbf{Speaking Rate} (4 features): Words-Per-Minute (WPM) were calculated for each turn the subject spoke, then the mean, minimum, maximum, and median were calculated across all segments for each subject.\\
\item \textbf{Questions} (2 features): A count of the question mark symbol `?' for each turn the subject spoke was calculated. Then the mean and cumulative sum across all segments were taken, for each subject. \\
\item \textbf{Hesitation} (2 features): A count of the transcription tag $<$um$>$ for each turn the subject spoke was calculated. Then the mean and cumulative sum across all segments were taken, for each subject.
\item \textbf{Vocabulary} (1 feature): The number of unique words expressed per subject during the exam were calculated.
\item \textbf{Out-of-Vocabulary (OOV) Rate} (1 feature): For each subject $s=i$, we identified the set of unique words $V_{s=i}$, and a list of unique words spoken by all other subjects ($V_{s \neq i}$). The OOV was computed as:
$$ \text{OOV} = \frac{|V^{C}_{s=i} \cap  V_{s \neq i}|}{|V_{s=i}|} $$
where $V^C$ denotes the complement of $V$.
\item \textbf{Language Perplexity (PPL)}: (2 features) For each subject $s=i$, a trigram model with Kneser-Ney discounting was trained on all other subjects. Using the trained model, the language perplexity was evaluated on each subject and all other subjects $s \neq i$ \cite{jurafsky2014speech}.  
$$ \text{PPL} = 2^H  \text{\,\,\,and\,\,\,} H =  -\frac{1}{M}\sum_{m=1}^{M} \text{log}\,\, p(w_m) $$
where $M$ is the size of the training vocabulary ($M = |V_{s \neq i}|$) of all subjects $s$ that aren't the $i^\text{th}$ subject, and $w_m$ is the $n^{\text{th}}$ word in the vocabulary. The SRILM Toolkit was used for this calculation \cite{stolcke2002srilm}.
\end{itemize}

\begin{figure}[b!]
  \centering
      \includegraphics[trim={0cm 0.5cm 0 0cm},clip,width=0.45\textwidth]{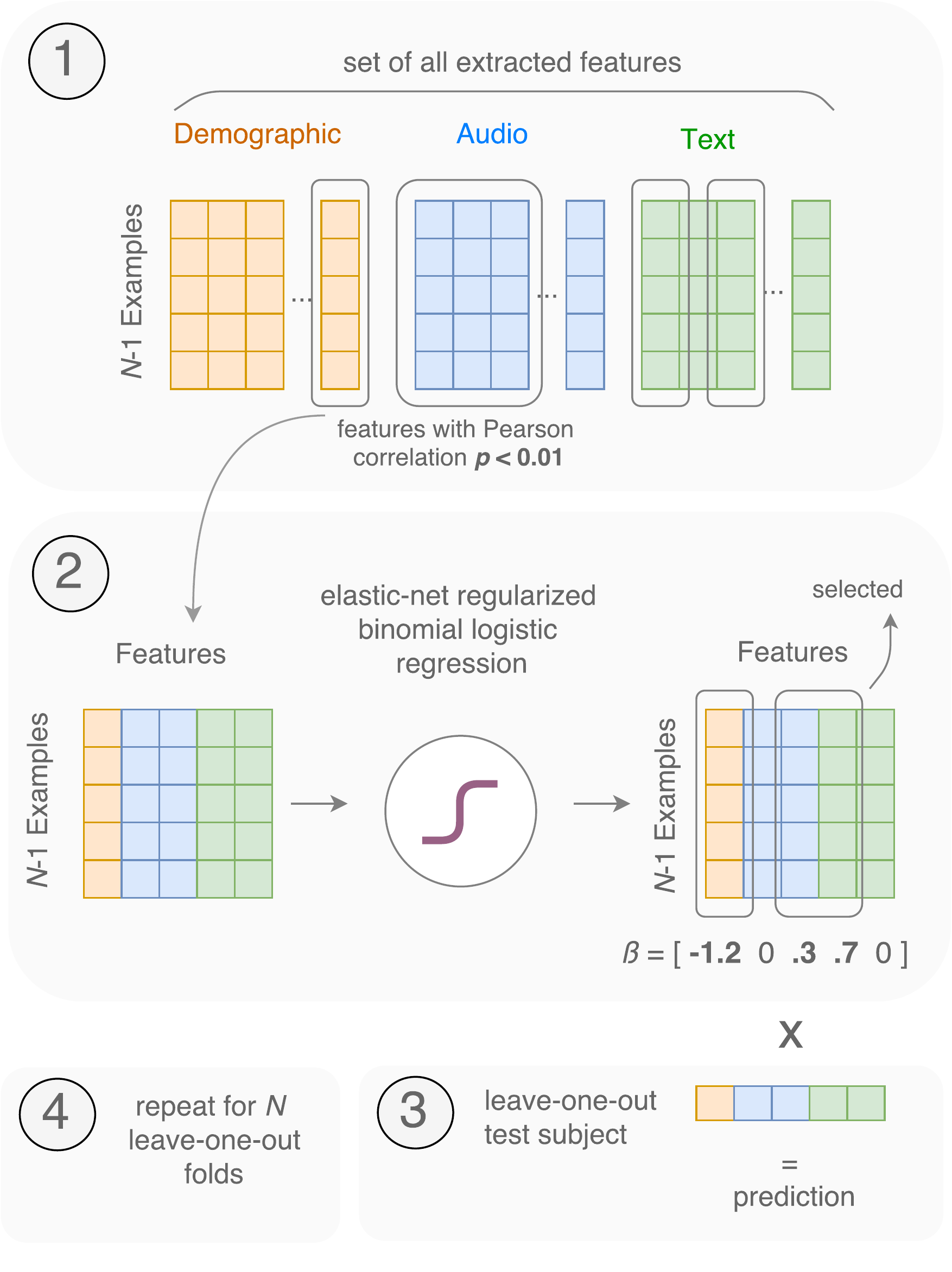}
  		\caption{\textbf{Modeling Technique and Feature Selection}. {\small (1) An initial set of features were selected if they had a statistically significant ($p$ $<$ 0.01) univariate Pearson correlation with the (training set, i.e. $N-1$ subjects) outcome. (2) An Elastic-net regularized binomial logistic regression is trained with these features, resulting in further feature selection. Features with non-zero model coefficients ($\beta$) were selected. (3) This model was evaluated on the held out leave-one-out test subject. (4) This training method was repeated for all leave-one-out ($N$) folds.}}
        \label{fig:modeling}
\end{figure}

\subsection{Model Choice and Evaluation Metrics}

Given the limited number of subjects with audio and text transcripts, and the importance of model interpretability, logistic regression was chosen as our modeling framework. We will describe each of the generated models in detail, below. 

To assess the performance of our models, the evaluation metrics we used were the Area Under the Receiver Operating Characteristic Curve (AUC), as well as the performance of the models at various points on the Curve: Accuracy, False Positive Rate (FPR), and True Positive Rate (TPR). We also performed the Hosmer-Lemeshow Test (HL-test) to evaluate the goodness of fit of the models \cite{hosmer1980goodness}. To evaluate the generalizability and robustness of our modeling techniques, we performed leave-one-out cross-validation. We also report the values of the model coefficients ($\beta$), and where applicable: odds ratio, confidence interval at 95\%, and the statistical significance of the features via the Wald Test.

\subsection{Baseline Models}

A baseline model using the 14 demographic features was trained on 6,258 subjects who had undergone neuropsychological examinations. Many subjects underwent multiple examinations over the years, which resulted in 12,258 training examples. We refer to this as the \textit{global} demographic model. 

A second baseline model using only demographic features was trained using data from the 92 subjects with audio and text transcriptions. This model used 9 of our 14 features (5 features were excluded due to zero variance). We refer to this as the \textit{local} demographic model. 

\subsection{Proposed Model}
In Figure \ref{fig:modeling} we show a visual representation of the proposed modeling pipeline, which we also describe here. While there were enough degrees of freedom to fit the \textit{local} demographic model (9 features and 92 examples), our dataset of 92 subjects was too small to accommodate the addition of the 251 text and acoustic features. We addressed the large feature to sample ratio in two steps: (1) an initial subset of features were selected that had statistically significant ($p$ $<$ 0.01) univariate Pearson correlations with the training set outcomes in each fold, (2) we then provided the subset of correlated features to an Elastic-net regularized binomial logistic regression model. In addition to the standard approach of minimizing the difference between predicted and true outcomes (i.e. deviance for binomial logistic regression), Elastic-net also minimizes the linear combination of $L1$ and $L2$ penalties of the estimated coefficients, which has the effect of producing sparse model coefficients thus implicitly selecting features via the resulting non-zero model coefficients. The objective function for an Elastic-net regularized binomial logistic regression model is defined as:

\begin{equation}
	\min_{\beta} \bigg[ DEV(\hat\beta) + \lambda\Big( (1-\alpha)||\beta||_{2}^2 + \alpha||\beta||_1 \Big)\bigg]
\end{equation}

where $\beta$ are the model coefficients being estimated, $DEV(\hat\beta)$ is the standard binomial logistic regression model objective function, $\lambda$ scales the influence of the regularization term, while $\alpha$  ranges between 0 and 1, and allows for a combination of $L1$ and $L2$ penalties. Thus when $\alpha$ = 0, only the $L2$ penalty forms the regularization term, and when $\alpha$ = 1, only the $L1$ penalty term appears. A more complete discussion of Elastic-net is described by \textit{Zou} et al. \cite{zou2005regularization}. 

We used the MATLAB implementation of the Elastic-net method, GLMNET \cite{hastie2014glmnet}, cross-validated over the training set, with the following parameters, $\alpha$ = $[0,1]$ with a step-size of 0.01, the cross-validated \textit{loss} set to `deviance', and the \textit{number-of-folds} over the training set equal to the number of subjects ($N$ - 1 = 91 examples). Up to 100 $\lambda$ values were automatically explored by GLMNET until a minimum deviance error threshold was reached. For our evaluation, $\lambda$ that was 1 standard error from the mean minimum error value across the cross-validated folds was used. All test performance values reported were according to the top performing model on the training set.


\section{Results} \label{sec:results}

\subsection{Demographic Model Coefficients}
Table \ref{table:odds} displays the demographic model coefficients ($\beta$), odds ratio, 95\% confidence interval and $p$-values for the \textit{global} and \textit{local} demographic models. For the \textit{global} model, as expected, age was positively associated with cognitive impairment ($p <$ 0.001). Similarly, as is also-well documented, an increasing level of education with at least some years of college (relative to some years of high school) was more negatively associated with cognitive impairment ($p$ $<$ 0.001). Employment status of retired, unemployed, and unemployed due to disability (relative to full-time employment) was positively associated with cognitive impairment ($p$ $<$ 0.05). High school degree, other categories of employment, and sex did not have a statistically significant association with outcome. In the \textit{local} model, age was again positively associated with cognitive impairment ($p <$ 0.05), while a high school degree and some college (relative to some years of high school) were negatively associated with cognitive impairment ($p$ $<$ 0.05). A college degree, employment status, and sex did not have a statistically significant association with the outcome.

\begin{table}[t!]
\centering
\caption{\textbf{Model Coefficients.} {\small Logistic regression model coefficients using \textit{global} and \textit{local} demographic features.}}
\label{table:odds}
\footnotesize
\begin{tabular}{lcccc}
\vspace{-0.3cm} & & & & \\

\multicolumn{5}{c}{GLOBAL DEMOGRAPHIC MODEL \textit{(N = 6,258)}} \\ \hline \hline
\textbf{Features}            
& \textbf{$\beta$} 
& \textbf{\begin{tabular}[c]{@{}c@{}}Odds Ratio \\$(e^\beta)$ \end{tabular}} 
& \textbf{95\% CI} 
& \textbf{$p$-val}  \\ \hline

\textbf{Age}          		  &2.02& 7.55        & [6.47, 8.81]  		& $<$ 0.001 \\                       
\multicolumn{5}{l}{\textbf{Education} (w.r.t some high school)}				         \\
\,\,\,\,high school     	  &-0.16& 0.85       & [0.68, 1.07]  		& 0.16 \\
\,\,\,\,some college          &-0.48& 0.62       & [0.48, 0.80]  		& $<$ 0.001 \\
\,\,\,\,college          	  &-0.59& 0.55       & [0.43, 0.72]  		& $<$ 0.001 \\

\multicolumn{5}{l}{\textbf{Employment} (w.r.t full-time)}				    \\
\,\,\,\,part-time             &0.07& 1.08       & [0.43, \,\,2.68]     & 0.87 \\
\,\,\,\,retired               &1.42& 4.16       & [2.00, \,\,8.64]     & $<$ 0.001 \\
\,\,\,\,unemployed            &1.66& 5.27       & [2.54, 10.94]        & $<$ 0.001 \\
\,\,\,\,disability    		  &1.68& 5.34       & [1.02, 28.13]        & $<$ 0.05 \\
\,\,\,\,never                 &0.88& 2.41       & [0.63,  \,\,9.25]    & 0.20 \\
\,\,\,\,volunteer             &-0.21& 0.81      & [0.23,  \,\,\,2.82]  & 0.74 \\
\,\,\,\,student     		  &-94.47& 0.00     & [0.00, \, \,\,\,-\,\,\, ]       & 1.00 \\
\,\,\,\,homemaker             &-94.47& 0.00     & [0.00, \, \,\,\,-\,\,\, ]       & 1.00 \\
\,\,\,\,other                 &0.64& 1.90       & [0.21, 17.65]        & 0.57 \\

\multicolumn{5}{l}{\textbf{Sex} (w.r.t male)} \\
\,\,\,\,female &-0.05 & 0.95     & [0.81, 1.12]        & 0.57 \\
\hline

\vspace{-0.1cm} & & & & \\

\multicolumn{5}{c}{LOCAL DEMOGRAPHIC MODEL \textit{(N = 92)}}                           \\ \hline \hline
\textbf{Features}           
& \textbf{$\beta$} 
& \textbf{\begin{tabular}[c]{@{}c@{}}Odds Ratio \\$(e^\beta)$ \end{tabular}} 
& \textbf{95\% CI} 
& \textbf{$p$-val}  \\ \hline

\textbf{Age}          		  & 1.43 & 4.20       & [1.18, 14.87]  		& $<$ 0.05 \\

\multicolumn{5}{l}{\textbf{Education} (w.r.t some high school)}				         \\
\,\,\,\,high school          & -3.44 & 0.03       & [0.00, 0.45]  		& $<$ 0.05 \\
\,\,\,\,some college         & -4.30 & 0.01       & [0.00, 0.28]  		& $<$ 0.01 \\
\,\,\,\,college              & -2.20 & 0.11       & [0.01, 1.24]  		& 0.08 \\

\multicolumn{5}{l}{\textbf{Employment} (w.r.t full-time)}				    \\
\,\,\,\,part-time             & -100.4 & 0.00     & [0.00, \,\,\,\,\,-\,\,\,\, ]        & 1.00 \\
\,\,\,\,retired               & 0.40   & 1.49     & [0.11, 20.13]     & 0.76 \\
\,\,\,\,unemployed            & -100.6 & 0.00     & [0.00, \,\,\,\,\,-\,\,\,\, ]        & 1.00 \\
\,\,\,\,volunteer             & 2.31   & 10.05    & [0.19,  531.93]   & 0.25 \\

\multicolumn{5}{l}{\textbf{Sex} (w.r.t male)} \\
\,\,\,\,female 		  		&-0.01  & 0.99      & [0.25, 4.03]      & 0.99 \\ \hline

\vspace{-0.6cm}
\end{tabular}
\end{table}

\begin{table*}[t!]
\centering
\footnotesize
\caption{\textbf{Results}. Performance of all models and features sets.}
\label{table:results}
\begin{tabular}{lcccccccccc}

 \multicolumn{10}{c}{\vspace{-0.2cm}} \\
 \multicolumn{10}{c}{BINOMIAL LOGISTIC REGRESSION MODEL (baseline)}                           \\ \hline \hline

\textbf{Features}                                    
& \textbf{AUC}
& \textbf{Acc.}
& \textbf{\begin{tabular}[c]{@{}c@{}} TPR @ \\ FPR10\% \end{tabular}}  
& \textbf{\begin{tabular}[c]{@{}c@{}} TPR @ \\ FPR5\% \end{tabular}}  
& \textbf{\begin{tabular}[c]{@{}c@{}} TPR @ \\ FPR0\% \end{tabular}}  
& \textbf{\begin{tabular}[c]{@{}c@{}} FPR @ \\ TPR90\% \end{tabular}} 
& \textbf{\begin{tabular}[c]{@{}c@{}} FPR @ \\ TPR95\% \end{tabular}} 
& \textbf{\begin{tabular}[c]{@{}c@{}} FPR @ \\ TPR100\% \end{tabular}} 
& \textbf{HL-test}  
& \textbf{$\alpha$} \\\hline

Generic - No Impairment  & 0.5 & 77 & 0 & 0 & 0 & 1 & 1 & 1 & - & - \\
Demographic - Local    & 0.74 &   72  &  0.56 & 	0.44 &	0    &	1    &	1    &	1    & 0 & -  \\  
Demographic - Global   & 0.79 &   83  &  0.14 & 	0.14 &	0.14 &	0.32 &	0.51 &	0.51 & 0 & -  \\ \hline

&&&&&&&&& \\

\multicolumn{10}{c}{BINOMIAL LOGISTIC REGRESSION MODEL - ELASTIC-NET REGULARIZED}                           \\ \hline \hline
\textbf{Features}                                    
& \textbf{AUC}
& \textbf{Acc.}
& \textbf{\begin{tabular}[c]{@{}c@{}} TPR @ \\ FPR10\% \end{tabular}}  
& \textbf{\begin{tabular}[c]{@{}c@{}} TPR @ \\ FPR5\% \end{tabular}}  
& \textbf{\begin{tabular}[c]{@{}c@{}} TPR @ \\ FPR0\% \end{tabular}}  
& \textbf{\begin{tabular}[c]{@{}c@{}} FPR @ \\ TPR90\% \end{tabular}} 
& \textbf{\begin{tabular}[c]{@{}c@{}} FPR @ \\ TPR95\% \end{tabular}} 
& \textbf{\begin{tabular}[c]{@{}c@{}} FPR @ \\ TPR100\% \end{tabular}} 
& \textbf{HL-test}  
& \textbf{$\alpha$}\\\hline

\multicolumn{1}{l}{{Text}}                   	& 0.69                       & 67                & 0.38                 & 0.24                & 0.14                & 0.59                 & 0.99                 & 0.99                  & $>$ 0.05 & 0.12 \\
\multicolumn{1}{l}{{Dem. + Text}}         		& 0.73                       & 74                & 0.33                 & 0.24                & 0.10                 & 0.56                 & 0.92                 & 0.93                  & $>$ 0.05 & 0.12 \\
\multicolumn{1}{l}{{Audio}}                  	& 0.90                       & 84                & 0.71                 & 0.48                & 0.14                & 0.25                 & 0.32                 & 0.38                  & $>$ 0.05 & 1.00\\
\multicolumn{1}{l}{{Dem. + Audio}}        		& 0.90                       & 84                & 0.67                 & 0.48                & 0.14                & 0.28                 & 0.31                 & 0.34                  & $>$ 0.05 & 1.00\\
\multicolumn{1}{l}{{Audio + Text}}           	& 0.92                       & 89                & 0.76                 & 0.62                & 0.29                & 0.17                 & 0.27                 & 0.44                  & $>$ 0.05 & 1.00\\
\multicolumn{1}{l}{{Dem. + Text + Audio}} 		& 0.92                    	& 89                & 0.76                 & 0.62                 & 0.38               & 0.17                 & 0.30                 & 0.44                   & $>$ 0.05 & 0.99          \\ \hline
\multicolumn{10}{l}{{AUC: Area Under the Receiver Operating Characteristic Curve. Acc: Accuracy. TPR: True Positive Rate. FPR: False Positive Rate.}}  \\ 
\multicolumn{10}{l}{{HL: Hosmer-Lemeshow. An HL-test greater than 0.05 indicates a well calibrated model.}}  \\ 
\vspace{-0.7cm}
\end{tabular}
\end{table*}


\subsection{Speech and Language Features}
Table \ref{table:results} shows the performance of the baseline models, as well as the performance of the Elastic-net regularized binomial logistic regression models for different combinations of speech and language features. 

For reference, a model that consistently guesses cognitive impairment (\textit{trivial} model) had an AUC of 0.5 (random) and an accuracy of 77\%. Using demographic features from the 92 subject subset (i.e. \textit{local} model), the AUC was 0.74. Exposing the demographic-based model to a larger group of subjects (i.e. \textit{global} model), and evaluating on the 92 subset, increased performance to an AUC of 0.79. This model also had a higher accuracy (83\%), and an improved TPR and FPR. Neither of the baseline demographic models were well calibrated according to the HL-test.

For the Elastic-net regularized models, an approach using text features alone did not yield a higher performing system (0.69 AUC) than the baselines. However, using audio based features resulted in higher performance (0.90 AUC) than the baseline models. Combining audio and text features resulted in the best performing model (0.92 AUC), while introducing demographic features along with audio, and audio-text combined did not improve performance with respect to AUC. Combining all three feature sets resulted in the best TPR at FPR of 0\%, but also increased FPR at TPR of 95\%. All Elastic-net regularized models were well calibrated according to the HL-test ($>$ 0.05)\footnote{Results of all modeling frameworks: \url{https://groups.csail.mit.edu/sls/publications/2017/ASRU17_alhanai_SI.pdf} }.

\subsection{Selected Features}
In Table \ref{table:coeff-elasticnet} we show the demographic, audio and text features selected by Elastic-Net. The values of the model coefficients shown in the table resulted from training the elastic net on all 92 subjects. We do not report confidence intervals and hypothesis testing because sparse estimators such as Elastic-net are difficult to interpret in the same way as a standard logistic regression model \cite{dezeure2015high}. Two text features were selected: the mean duration of each subject's turn (Segment Duration), as well as the number of `?' symbols (Question Marks) transcribed in the text. The segment duration had a negative association with the outcome (cognitive impairment), while the number of Question Marks had a positive association with the outcome. From the audio features, pitch based measures; minimum pitch and standard deviation of the jitter were selected. Both features had negative associations with the outcome. The rest of the features selected were energy based (MFCC 2 (sd), 3 (max), 6 (median), 10 (sd), 13 (min)) and included the difference between frames (MFCC 3 (mean, median), 8 (median)). All energy based features had a mix of negative and positive associations with the outcome. 

\section{Discussion} \label{sec:discussion}


The superior performance of audio features indicated that there was low level information about the speaker's pitch and its variance (jitter) that was predictive of cognitive impairment. These findings are in line with the literature \cite{meilan2014speech, horley2010emotional}, where decreasing pitch and variance may indicate less expressive speech. We note that different forms of dementia (e.g. Alzheimer's, Vascular) and co-occurring conditions (e.g. Parkinson's Disease) may exhibit differing speech pathologies, as observed by \textit{Illes} \cite{illes1989neurolinguistic}. Therefore we interpret our results to be capturing a broad spectrum of cognitive disorders. 

Although text features alone were not found to be predictive of the outcome, some were found to be meaningful when combined with audio features. Features that captured hesitation (via counts of the question mark symbol `?'), and shorter time taken to respond to the question (via mean Segment Duration) indicated the subject's struggle in responding fully and/or recalling words, agreeing with observations in the literature \cite{lopez2015automatic}. However, features capturing syntax (vocabulary size, OOV rate) and coherence in speech (perplexity) were not selected. It may be that such features are more useful when modeling with a larger dataset, with previous studies indicating that early predictors of cognitive impairment (let alone onset) may be observed at the syntactic and semantic level of speech, more so than with acoustics \cite{taler2008language}. 

Although demographic features performed well (0.79 AUC), they were not necessary when combined with audio and text features as evidenced by their absence from the Elastic-net regularization. This indicates that text and audio features were capturing at least equivalent information such as age and gender of the speaker (via pitch and energy), and/or were capturing information that was even more predictive of cognitive impairment, overshadowing the information content of demographic features. These results suggest that cognitive impairment can be screened for without having any information on the subject's profile, using audio recordings alone, and without the constraint of medical visit logistics, missing medical history, or sparse medical evaluations.


The results of modeling with audio and text features indicates that this source of information allows for models that are not only high-performing, but also well calibrated. Importantly, real-world diagnostic systems requires that predictions made by models are well calibrated. That is, if a model assigns a 30\% probability of cognitive impairment, it is important that cognitive impairment occur 30\% of the time. Well calibrated systems allow clinicians and families to make informed judgments about risk. 


\section{CONCLUSIONS} \label{sec:conclusions}

In this study we utilized audio recordings of 92 subjects undergoing neuropsychological examinations at the Framingham Heart Study. We found that combining audio and text features provided the best performance in detecting cognitive impairment (0.92 AUC), and was superior to the baseline approaches that used demographic features from the 6,258 subject cohort (0.79 AUC). Given the high-dimensionality of the feature set (265 total), we used an Elastic-net binomial logistic regression model which selected 12 features. We found that decreasing pitch, decreasing jitter, shorter speech segment lengths, and and an increasing number of questions by the subject were positively associated with cognitive impairment. Our methodology did not explicitly model the structure and components of the neuropsychological exams subjects underwent, which allows for it to generalize to other scenarios such as informal spoken interactions. 

\begin{table}[h]
\centering
\footnotesize
\caption{\textbf{Selected Features.} {\small Features selected from demographic, audio, and text feature set using Elastic-net regularization (0.92 AUC, $\alpha$ = 0.99, $\lambda$ = `lambda\_1se'). The \% Selected indicates the proportion of leave-one-out folds the feature was selected.}}
\label{table:coeff-elasticnet}
\begin{tabular}{lllcc}
\\
\hline \hline

\textbf{Features} &
 &
\textbf{Type} &
$\beta$ & 
\begin{tabular}[c]{@{}c@{}} \% \textbf{Selected} \end{tabular} \\ \hline

 MFCC 13 &(min) 			 & Audio & 0.0043 & 60 \\
 Segment Duration &(mean) & Text & -0.0083 & 82  \\
 MFCC 10 &(sd) 			 & Audio & -0.1346 & 89 \\
 MFCC 2 &(sd) 			 & Audio & -0.2514 & 98 \\
 MFCC 3 diff. &(mean)     & Audio & -0.1526  & 99 \\
 Question Mark &(sum)     & Text  &  0.0171  & 100  \\
 MFCC 6 &(median) 		 & Audio &  0.1741 & 100 \\
 Pitch &(min)   			 & Audio & -0.2430 & 100 \\ 
 MFCC 8 diff. &(median)   & Audio & 0.4187 & 100 \\
 Jitter &(sd) 		     & Audio & -0.5337 & 100 \\
 MFCC 3 &(max)    		 & Audio & -0.6168  & 100  \\
 MFCC 3 diff. &(median)   & Audio & -0.7620 & 100 \\
 \hline

\end{tabular}
\end{table}

\newpage 
\small
\label{sec:references}

\bibliographystyle{IEEEbib}
\bibliography{strings,refs}

\end{document}